\documentclass[10pt,twocolumn,letterpaper]{article}

\usepackage{iccv}              
\usepackage{multirow}
\newcommand{\tworow}[2]{\begin{tabular}[c]{@{}c@{}}#1\vspace{-2pt}\\#2\end{tabular}}

\usepackage[accsupp]{axessibility}
\definecolor{iccvblue}{rgb}{0.21,0.49,0.74}
\usepackage[pagebackref,breaklinks,colorlinks,allcolors=iccvblue]{hyperref}

\def\paperID{6125} 
\def\confName{ICCV}
\def\confYear{2025}

\title{Video Color Grading via Look-Up Table Generation}

\author{Seunghyun Shin\textsuperscript{\rm 1} \qquad %
Dongmin Shin\textsuperscript{\rm 2} \qquad %
Jisu Shin\textsuperscript{\rm 1} \qquad %
Hae-Gon Jeon\textsuperscript{\rm 2}\footnotemark[2] \qquad %
Joon-Young Lee\textsuperscript{\rm 3}\footnotemark[2]\\
\textsuperscript{\rm 1}GIST \qquad \textsuperscript{\rm 2}Yonsei University\qquad \textsuperscript{\rm 3}Adobe Research\\
}

\begin{document}
\twocolumn[{%
\maketitle
\vspace{-11mm}
\renewcommand\twocolumn[1][]{#1}%
\begin{center}
    \centering
    \captionsetup{type=figure}
    \includegraphics[width=\textwidth]{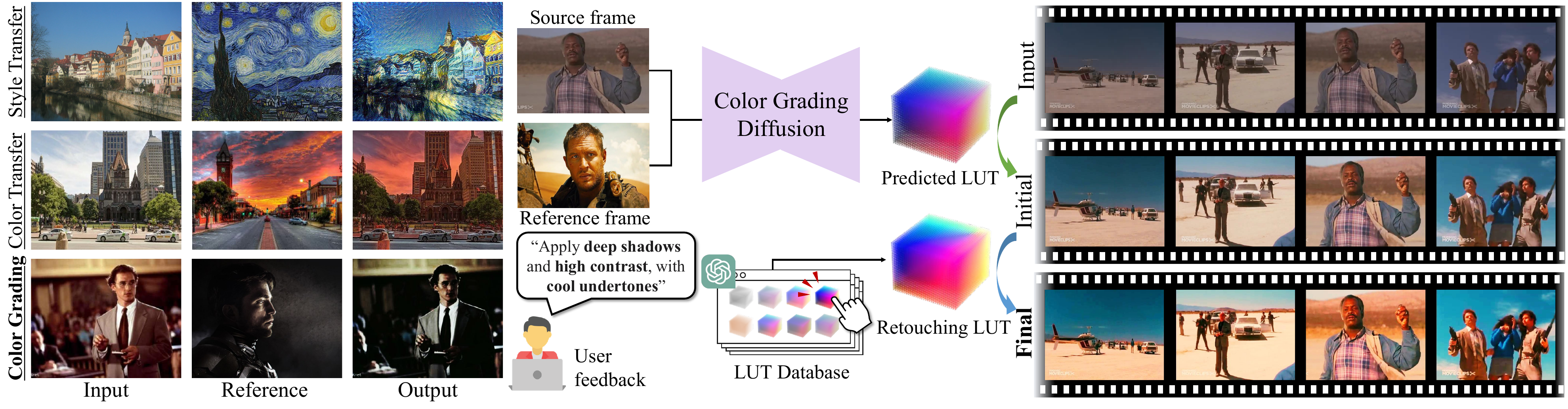}
    \vspace{-7.5mm}
    \captionof{figure}{(Left) Difference between conventional style/color transfer and the color grading task. Conventional color/style transfer aims to match colors/textures of reference images with the input image, while video color grading captures the reference frame's artistic intents. 
    (Middle) An overview of the proposed framework for reference-based video color grading and user-preferred retouching via a text prompt.
    (Right) An example video clip from our video color grading framework based on the generated LUTs. }
    \label{fig:teaser}

\end{center}%
}]
\begin{abstract}
Different from color correction and transfer, color grading involves adjusting colors for artistic or storytelling purposes in a video, which is used to establish a specific look or mood.
However, due to the complexity of the process and the need for specialized editing skills, video color grading remains primarily the domain of professional colorists. 
In this paper, we present a reference-based video color grading framework.
Our key idea is explicitly generating a look-up table (LUT) for color attribute alignment between reference scenes and input video via a diffusion model. 
As a training objective, we enforce that high-level features of the reference scenes like look, mood, and emotion should be similar to that of the input video.
Our LUT-based approach allows for color grading without any loss of structural details in the whole video frames as well as achieving fast inference. We further build a pipeline to incorporate a user-preference via text prompts for low-level feature enhancement such as contrast and brightness, etc. Experimental results, including extensive user studies, demonstrate the effectiveness of our approach for video color grading. Codes are publicly available at https://github.com/seunghyuns98/VideoColorGrading.

\renewcommand*{\thefootnote}{\fnsymbol{footnote}}
\footnotetext[2]{Corresponding Author}
\renewcommand*{\thefootnote}{\arabic{footnote}}

\end{abstract}

\section{Introduction}
\label{sec:intro}

In the film ``Parasite", color grading is used to express class differences: the wealthy family's home is rendered in warm, vibrant tones, evoking a sense of comfort and wealth, while the poor family's environment is rendered in cool, desaturated tones, emphasizing their hardship and separation from privilege~\cite{Parasite}. Color grading aims to add depth and meaning to each scene, guiding the viewer's perception and interpretation of a story. It is a post-production process for video editing in the film industry, and consists of adjusting color and contrast to serve a creative or stylistic choice.
However, because color grading requires a blend of technical skills and artistic vision, including a deep understanding of color theory, visual storytelling, and mood setting, this process remains largely within the realm of professional colorists. 

Until now, studies on color grading have been conducted in both computer vision and graphics fields. However, they define the color grading task as color distribution matching between input and reference images~\cite{lee2000color, pitie2007automated, bonneel2013example}. 
Unfortunately, these methods are insufficient for many
practical purposes.
Since color grading is a process of manipulating the colors in videos for artistic or stylistic reasons, it requires the transfer of high-level features, such as atmosphere, tone and vibe, beyond simply adjusting color distribution. In addition, users want to have various creative controls over the initial products.

In this paper, we propose a novel video color grading framework that allows users to generate aesthetically pleasing videos based on preferred references in~\cref{fig:teaser}. Our entire pipeline is inspired by the workflow of professional colorists. Their typical process involves selecting a key-frame and then adjusting the frame to mimic their subjective characteristics that entangle contrast, tone, style and mood in the image. They then apply these adjustments consistently to all frames, and finally make minor retouching adjustments to all frames to match their aesthetic preferences~\cite{Kennel2006, Charles2019}.

Following this workflow, we begin by selecting one of input frames as a target frame that closely resembles a reference image with respect to scene configuration and appearance, based on CLIP~\cite{radford2021learning} feature similarity. When using video reference, we similarly choose the best matching frame. 
We then extract high-level visual features to account for the subjective characteristics from the reference image. To do this, we train a Grading Style Extraction network (GS-Extractor). After that, we generate a look-up table (LUT) to transfer the color attributes to the input video. 

To ensure strong generalization and diverse color representation, we explicitly generate a LUT via a diffusion model (named L-Diffuser). Once the LUT is generated, it is applied across all frames in the input video, achieving temporal consistency and fast adjustments without any artifact. As a last step of our framework, we facilitate additional edits to the initial color graded video by user's text prompts, such as ``increasing contrast" or ``emphasizing red tones".

We demonstrate the effectiveness of our pipeline, showing superior performance over relevant reference-based retouching methods in terms of both qualitative and quantitative evaluations. In addition, extensive user studies further validate our framework, capturing aesthetic qualities and subtle details that are challenging to measure solely through computational metrics.

\section{Related Works}
\subsection{Reference-based Image Retouching}
Photorealistic style transfer involves adapting the color characteristics of a reference image to a target image while preserving the target’s structural integrity and fine details, unlike style transfer~\cite{chen2017stylebank, cheng2021style, deng2022stytr2, dumoulin2016learned, gatys2016image, hong2021domain, huang2017arbitrary, johnson2016perceptual, li2016combining, li2017diversified} that typically focuses on transferring both texture and color.
Traditional methods~\cite{reinhard2001color, pitie2005n, pitie2007automated, welsh2002transferring} achieve photorealistic style transfer by aligning statistics for low-level features, like the mean, variance or histogram of filter responses, between the reference and target images. 
Recently, deep learning frameworks have been used to capture complex color distributions and richer image content representations, enabling more detailed feature mapping between images~\cite{luan2017deep, cheng2021style, li2019learning, chiu2022pca, wen2023cap}.
For example, several studies~\cite{chiu2022photowct2, li2018closed, yoo2019photorealistic} leverage wavelet-pooling to preserve high-frequency details while transferring low-frequency information using whitening and color transformation (WCT). To preserve structure information and enable fast style transfer, PhotoNAS~\cite{an2020ultrafast} utilizes skip connections in multi-layers and adopts an iterative pruning to compress networks. Histo-GAN~\cite{afifi2021histogan} utilizes a generative approach capable of transforming input distributions into target domains conditioned on given references. Specifically, it leverages the color histogram of a reference image as a conditioning vector to effectively align the input image's color distribution with that of the reference. CAP-VSTNet~\cite{wen2023cap} reduces a loss of structural details in the style transfer task by adopting reversible networks~\cite{an2021artflow}. This network structure allows feature maps from earlier layers to be reconstructed from deeper layers where transfer operations are performed. However, the issue of maintaining content consistency in the output image has not yet been fully resolved, as the distribution matching in latent spaces often results in a loss of fine details from the original images.

\begin{figure*} [h]
    \centering
    \includegraphics[clip, width=\textwidth]{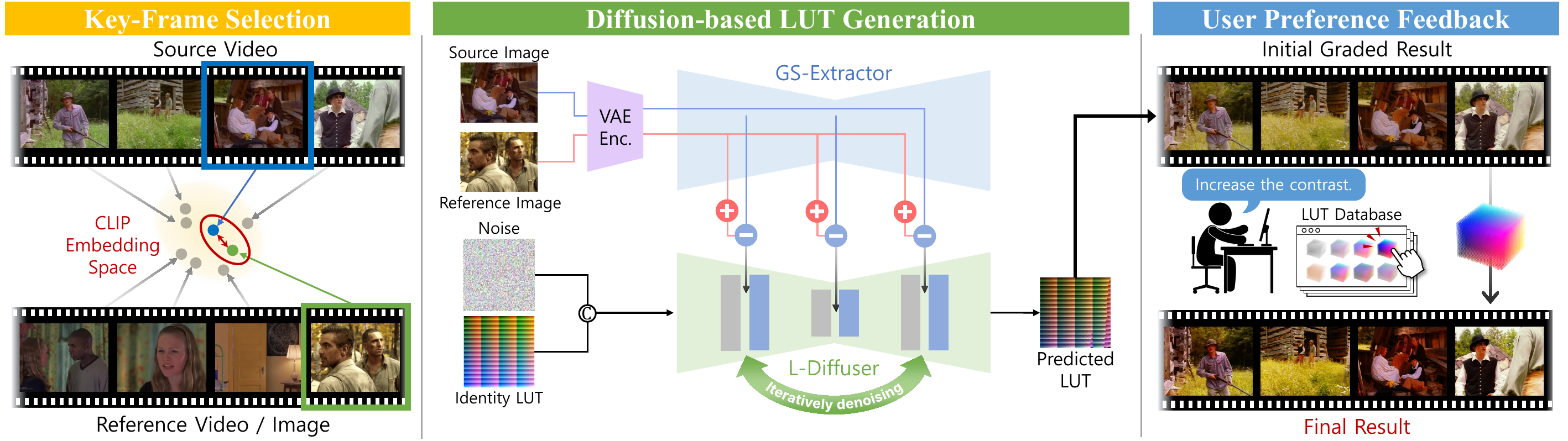}
    \vspace{-7.5mm}
    \caption{An overview of our video color grading framework: (1) key-frame selection; (2) LUT generation; (3) user preference feedback process via text prompts.}
    \vspace{-5mm}
    \label{fig:overview}
\end{figure*}
\subsection{Video Color Retouching}
In the realm of video style transfer, where image style transfer methods are extended to videos, such as movie clips or temporally calibrated image sequences, a primary challenge lies in maintaining temporal coherence across frames while preserving stylized content details. Previous works~\cite{chen2017coherent, deng2021arbitrary, huang2017real, ruder2018artistic, xia2021real, li2019learning, wu2020preserving, gao2020fast, wang2020consistent} have addressed this issue by devising constraints to prevent severe color variations between frames and to mitigate flickering artifacts. These constraints typically involve designing loss terms for the temporal coherence between consecutive frames or coupling style and content features across the video sequence. Optical flow is often employed to warp stylized frames directly onto subsequent frames, enforcing inter-frame consistency~\cite{xia2021real, ruder2018artistic, chen2017coherent, gao2020fast}. However, due to its inherent computational time complexity, some studies have tackled this issue at the feature level by applying linear transformations between content and style features, integrating self-attention mechanisms or using similarity loss for regularization during training~\cite{wu2020preserving, li2019learning, deng2021arbitrary, wang2020consistent, wu2022ccpl}. 
For example, CCPL~\cite{wu2022ccpl} extends photorealistic style transfer to video level by applying a contrastive loss between local patches of the adjacent frames to reduce temporal inconsistency. Recently, UniST~\cite{gu2023two} proposes a unified image-video joint learning framework for style transfer, which leverages mutual benefits between image and video domains. Nonetheless, they are still unsuitable for color grading because variations inevitably arise as they process each frame individually, which leads to the lower quality of the original video.

\subsection{Deterministic Color Mapping} 
Deterministic color mapping is devised to avoid artifacts that often occur in the latent-space-based image retouching. A work in~\cite{ho2021deep} predicts image-adaptive parameters of pre-defined filters to control brightness and contrast of input images. Works in~\cite{xia2020joint, shin2024close} generate bilateral grids to transfer colors or styles of reference images to input images in a spatial-aware manner. The other~\cite{ke2023neural} utilizes MLP layers to transform input images for efficient color transfer.

Meanwhile, LUT-based color adjustments have long been adopted due to their flexibility of color transformations in commercial products, like `Premiere Pro' or `Photoshop'. Recently, image-adaptive LUT predictions have been widely studied for image enhancement and photorealistic style transfer~\cite{zeng2020learning, liu20234d, conde2024nilut, zhang2022clut, yang2022seplut, yang2022adaint, wang2021real, chen2023nlut, lin2023adacm} as well.
Works in \cite{zeng2020learning, wang2021real} generate LUT bases and their weights, and some methods \cite{yang2022adaint, wang2021real, zhang2022clut} focus on LUT bases compressions while retaining their strong mapping capability. 

Obviously, LUT prediction has been applicable to color transfer tasks. NLUT~\cite{chen2023nlut} extends the concept of \cite{zhang2022clut} to the color transfer task by generating LUTs based on a pair of input and reference images. AdaCM~\cite{lin2023adacm} predicts MLP layer values, which transforms an identity LUT into an output LUT for color transfer. Unfortunately, they are typically constrained to specific style domains and require fine-tuning for unseen styles. They also suffer from color distortion artifacts when following standard protocol which applies predicted LUTs to whole video frames. In addition, since they exploit LUTs as intermediate operators like implicit representations, the generated LUTs are primarily biased toward only colors of input and reference frames.

Compared to the previous works, our method explicitly learns generalized LUTs by conditioning on input and reference image features. This enables our LUTs to accurately map the high-level features between input and reference frames and to robustly account for color relationships of unseen frames. As a result, our method yields a series of artifact-free color graded frames, allowing us to produce videos with temporal consistency.

\section{Method}
\label{sec:method}
In this section, we present a video color grading framework. The main challenge in this work is how to extract and transfer high-level features of either reference image or video, rather than simply color distribution, into an input video without any loss of visual contents.
We begin by describing our dataset construction for video color grading in Sec.~\ref{subsec:dataset}. We then introduce the proposed pipeline in Sec.~\ref{subsec:overview}. Lastly, training details are provided in Sec.~\ref{subsec:training strategy}. 
\subsection{Dataset Construction}

\label{subsec:dataset}
We create a new dataset because we could not find an established public benchmark for our video color grading task. We use the Condensed Movie Dataset~\cite{bain2020condensed}, which consists of over 33,000 clips from 3,600 movies covering the salient parts of the films and has two-minutes running time for each clip on average. The rationale for selecting this dataset is that movies obviously contain professional color grading effects and are useful as high-quality references in this work. Furthermore, we collect $100$ LUT bases which are selected as distinctive LUTs from the $400$ LUTs of the Video Harmonization Dataset~\cite{lu2022deep}. Since LUTs are able to ensure diverse outputs, it is flexible to edit videos with various characteristics. Either by directly applying or making new LUTs through interpolation and extrapolation, we can generate infinite number of LUTs and input-output pairs for training.
\subsection{Network Architecture}
\label{subsec:overview}

The overall pipeline of the proposed method is illustrated in \cref{fig:overview},
We start with a key-frame selection which finds a frame pair from input and reference videos. This frame pair is used to guide a diffusion model to generate explicit LUTs for transferring high-level features from the reference to the input video. After applying LUTs across entire video frames, user can adjust the initial results with their preferences through text descriptions.

\noindent\textbf{Key-Frame Selection}
Colorists tend to choose a reference style for video color grading in practice. The reference image or video has semantically similar scene configurations (i.e., subject, action, and angle of the shot) and high-level styles (i.e., atmosphere, emotion, and tone) that the user implicitly seeks. Based on this tendency, we utilize a pre-trained CLIP-Image encoder~\cite{radford2021learning} to find the most semantically similar frame pair from input and reference video. 

Let us assume the input and reference frames as $I_{1:M}$ and $I'_{1:N}$, respectively, where $M$ and $N$ mean the number of video frames. We find a key frame pair $(I_{\hat{m}}, I'_{\hat{n}})$ that maximizes a cosine similarity as follows:
\begin{equation}
    (\hat{m}, \hat{n}) = \underset{m,n}{\mathrm{argmax}}\frac{f_{m} \cdot f'_{n}} {\left\| f_m \right\| \left\| f'_{n} \right\|}, ~~~~ f^{(')}_x = \textit{CLIP}(I^{(')}_x),
    \label{eq:cossim}
\end{equation}
where $m \in \{1,...,M\}$ and $n \in \{1,...,N\}$ denote frame indices in each video. We sample $1$ frame per second since semantics remain stable across short intervals.

\noindent\textbf{GS-Extractor}
Our LUT generation consists of two main components. First, we propose a module to extract high-level features from frames. We leverage ReferenceNet architecture~\cite{hu2024animate} to utilize their effectiveness of extracting user-preferred features. ReferenceNet takes two different U-Nets, one for extracting features from a reference frame and the other for video generation which is conditioned by the features via spatial attention~\cite{hu2024animate, chang2023magicdance, xu2024magicanimate}. 

Similarly, we employ two distinct U-Nets which have identical structures: (1) \textit{GS-Extractor}, which extracts subjective features from a reference frame; (2) image editor, which adjusts an input frame to align its subjective characteristics with those of the reference frame through a denoising process. In a training phase, we choose a frame for the input~$(I_{\hat m})$ and the target~$(I'_{\hat m})$ frame in a movie, and apply a LUT to one of them. A reference frame~$(I'_{\hat n})$ is a different frame in the same movie. If the LUT is applied to the target frame, it is also used to the reference frame, so that they have the same color attributes and similar semantics. They are encoded into a compact latent representation~\cite{rombach2022high}: 
\begin{equation}
z^{I_{\hat m}}_0=\mathcal{E}(I_{\hat m}),~~~~ z^{I'_{\hat n}}_0=\mathcal{E}(I'_{\hat n}),~~~~ z^{I'_{\hat m}}_0=\mathcal{E}(I'_{\hat m}),
\label{eq:embedding}
\end{equation}
where $\mathcal{E}$ is a variational auto-encoder (VAE)~\cite{kingma2014, van2017neural}. 
The reference frame latent $z^{I'_{\hat n}}_0$ is then fed into the GS Extractor to produce a feature $G(z_0^{I'_{\hat n}})$. 
Motivated by diffusion models for image editing~\cite{brooks2023instructpix2pix, kawar2023imagic}, we use the feature from GS-Extractor as a conditioning vector to edit visual features of the input frame similar to that of the target frame.

The target frame latent $z^{I'_{\hat m}}_0$ is diffused in $K$ time-steps to produce noisy latent $z^{I'_{\hat m}}_K$ which is together used as input of the denoising U-Net with the source latent. The objective function for the denoising network $\epsilon_\theta$ is formulated as: 
\begin{equation}
\ell = \mathbb{E}_{z^{I'_{\hat m}}_k,z^{I_{\hat m}}_0,G(z^{I'_{\hat n}}_0),\epsilon,k}\Big[||\epsilon - \epsilon_\theta(z^{I'_{\hat m}}_k,z^{I_{\hat m}}_0,G(z^{I'_{\hat n}}_0),k)||^2_2\Big] ,
\label{eq:SD}
\end{equation}
where $\|\cdot\|_2$ is the $\ell_2$ distance and $k\in\{1, ..., K\}$ is a diffusion step which is uniformly sampled during training.

\noindent\textbf{L-Diffuser} Due to the use of the compressed latent space, high-frequency details may be lost during encoding. We propose a LUT generation model, named L-Diffuser. 
Here, we replace the image editor, only used for training GS-Extractor to extract color attributes, with L-Diffuser. 

We freeze GS-Extractor, and train L-Diffuser to generate LUTs from a random noise. Rather than directly creating LUTs, we train the model to produce variations from an identity LUT, which maps $(r,g,b)$ values to themselves.
Let's denote a difference between a LUT and the identity LUT as $\Delta L\in \mathbb{R}^{16\times 16\times 16 \times 3}$. We can reshape it as $\Delta L' \in \mathbb{R}^{64\times 64\times 3}$ to match the dimensions with the image latent. After diffusing it for $K$ time-steps, $\Delta L'_K$ is used as the model input. Furthermore, we concatenate it with the identity LUT ($L^I$) to make the basis of the LUT variations. 

To introduce a conditioning vector into L-Diffuser, we obtain a gradient direction vector by subtracting the source feature from the reference one through GS-Extractor. It allows us to align high-level features of the input and the reference image. 
The conditional vector construction is formulated as:
\begin{equation}
C = G(z^{I'_{\hat n}}_0)- G(z^{I_{\hat m}}_0).
\label{eq:direction}
\end{equation}
Finally, the objective function is defined as:
\begin{equation}
\ell = \mathbb{E}_{\Delta L'_k, L^I, C, \epsilon, k}||\Big[\epsilon - \epsilon_\theta(\Delta L'_k, L^I, C, k)||^2_2\Big].
\label{eq:SD2}
\end{equation}
By adding the identity LUT to the output of L-Diffuser, we obtain LUTs that transform subjective characteristics from the reference to the input image. It is applied through entire video frames to achieve its temporal coherency.

\begin{table*}[t]
    \centering
    \resizebox{\linewidth}{!}{%
    \begin{tabular}{c|c|ccccccc|c}
    \hline
    \hline
    Dataset & Method & WCT2~\cite{yoo2019photorealistic} & PhotoNas~\cite{an2020ultrafast} & Deep Preset~\cite{ho2021deep} &HistoGAN~\cite{afifi2021histogan} & CCPL~\cite{wu2022ccpl} & CAP-VST~\cite{wen2023cap}& NLUT~\cite{chen2023nlut} & ~~~~~\textbf{Ours}~~~~~ \\
    \hline
    \multirow{6}{*}{\tworow{Condensed}{ Movie}}&PSNR$\uparrow$ & 19.97 & 17.08 & 20.70 &18.36 & 17.13 & 20.55& \underline{21.41} & \textbf{24.55} \\
    &SSIM$\uparrow$ & 0.7490 & 0.6515& 0.7429 & 0.6891 & 0.6195 & \underline{0.7642} & 0.7398& \textbf{0.8445} \\
    &LPIPS$\downarrow$ & \underline{0.3026} & 0.3936& 0.3221 & 0.3595 & 0.3951 & 0.3034 & 0.3028 & \textbf{0.1457} \\
    &BRISQUE$\downarrow$ & 41.57 & 41.71&44.63 & 45.68 & \underline{40.40} & 40.56& 42.39 &\textbf{40.23} \\
    &Blur$\downarrow$ & 0.4207 & 0.4766&0.4893 & 0.4834 & 0.4297 & 0.4385 &\textbf{0.4134} &\underline{0.4183} \\
    &Inference Time~(s) & 130.08 & 54.48&25.64  & 50.47 & 38.55 & 46.97 & \underline{21.24}&\textbf{12.10} \\
    \hline
    \multirow{5}{*}{Adobe5k}&PSNR$\uparrow$ & 14.83 & 14.67 & 15.99& 15.20 & 15.81 & 16.33 &\underline{16.76} & \textbf{18.78} \\
    &SSIM$\uparrow$ & 0.6622 & 0.6701 & 0.7590 & 0.6970 & 0.7016 & \underline{0.7855} & 0.7298& \textbf{0.7966} \\
    &LPIPS$\downarrow$ & 0.3814 & 0.4010  &\underline{0.2744} & 0.3859 & 0.4560 & 0.3272 & 0.3915 & \textbf{0.2689}\\
    &BRISQUE$\downarrow$ & 27.97 & 23.98 & 28.19 & 41.27 & 29.75 & \textbf{16.08}& 21.03 &\underline{20.72} \\
    &Blur$\downarrow$ & 0.3841 & \underline{0.3409}&0.3719 & 0.4269 & 0.3851 & 0.3586 &0.3558 &\textbf{0.3360} \\
    \hline
    \hline
    \end{tabular}
    }\vspace{-3mm}
    \caption{Quantitative results on our dataset and public Adobe5k dataset. 
    (Best: \textbf{Bold}, Second best: \underline{Underline})}
    \vspace{-6mm}
    \label{tab:main_quant}
\end{table*}

\noindent\textbf{User Preference Feedback} As a final step, we build a simple yet useful procedure to reflect user's feedback for additional video retouching. In this procedure, users can give their feedback with text prompts. To do this, we designate a description for each LUT in our database. The description explains about potential outcomes when applying each LUT. We utilize GPT-4~\cite{achiam2023gpt} to initially generate the descriptions. Then, we manually verify and refine them to fix missing and wrongly labeled components. We embed these descriptions in vectors from the CLIP-Text encoder. When we receive an instruction from a user, we also embed it into a CLIP-Text space and calculate a cosine similarity based on the description vectors to find the best matching one. 

\subsection{Implementation Detail}
\label{subsec:training strategy}

To train our model, the input and reference frames are randomly selected from the training dataset. We divide LUTs with $90$ and $10$ LUTs for training and test, respectively. They are mixed through interpolation and extrapolation to generate new LUTs for the better robustness of the model. We initialize GS-Extractor with the pre-trained weights from Stable 
Diffusion~\cite{rombach2022high} and L-Diffuser with Gaussian weights. The weights of the VAE-encoder, CLIP image and text encoder are all fixed. We build our video color grading pipeline using PyTorch~\cite{paszke2019pytorch}, utilize the AdamW~\cite{loshchilov2017decoupled} optimizer with $\beta_1$=$0.9$ and $\beta_2$=$0.999$, and set a learning rate to $1e-5$. Experiments are conducted on 4 NVIDIA A100 GPUs. For training, frames are resized and center-cropped to a resolution of $512\times512$. We train GS-Extractor for 110K steps with a batch size of 16 and 100K steps with a batch size of 64 for L-Diffuser. During inference, we use a DDIM sampler for 25 denoising steps.

\section{Experimental Results}
\label{sec:exp}
\vspace{-1mm}
We conduct extensive experiments to demonstrate the effectiveness of our method. First, we provide details of the experimental setup in \cref{subsec:setup}.
In \cref{subsec:quantitative}, we explain the quantitative evaluation to numerically compare our method with state-of-the-art (SOTA) color/photorealistic style transfer methods. In~\cref{subsec:anal}, we analyze the proposed method with respect to the objective function and advantages of diffusion-based LUT generation. Moreover, we carry out an ablation study to validate the effectiveness of each component in our method in \cref{subsec:abl}. 
Lastly, particularly considering the special nature of video color grading task, we conduct a comprehensive user study to evaluate the transfer of subjective characteristics which are hard to measure through numeral metrics in \cref{subsec:userstudy}.

\vspace{-1mm}

\subsection{Experimental Setup}
\label{subsec:setup}
\vspace{-1mm}

We perform a quantitative evaluation on our processed dataset. $10$ LUTs are applied to a test split of Condensed Movie Dataset to generate Input/GT video pairs. Since the running time of the shortest video in the test split is 30 seconds, we conduct this experiment across 480 frames at 512$\times$512 resolutions and select the last frame in the video as a reference frame. We further validate our method using the public Adobe5k dataset, which contains 5,000 input images and five retouched versions from professional photographers. We use the test split of input images as sources, and the corresponding retouched images from each expert’s set as targets, while randomly selecting reference images from the same expert’s set. We compare our method with representative photorealistic style transfer methods which are publicly available: WCT2~\cite{yoo2019photorealistic}, PhotoNAS~\cite{an2020ultrafast}, 
Deep preset~\cite{ho2021deep}, HistoGAN~\cite{afifi2021histogan}, CCPL~\cite{wu2022ccpl}, CAP-VSTNet~\cite{wen2023cap} and NLUT~\cite{chen2023nlut}. To ensure a fair comparison, we use the official resources provided by authors.

\begin{figure*} [t]
    \centering
    \includegraphics[clip, width=\linewidth]{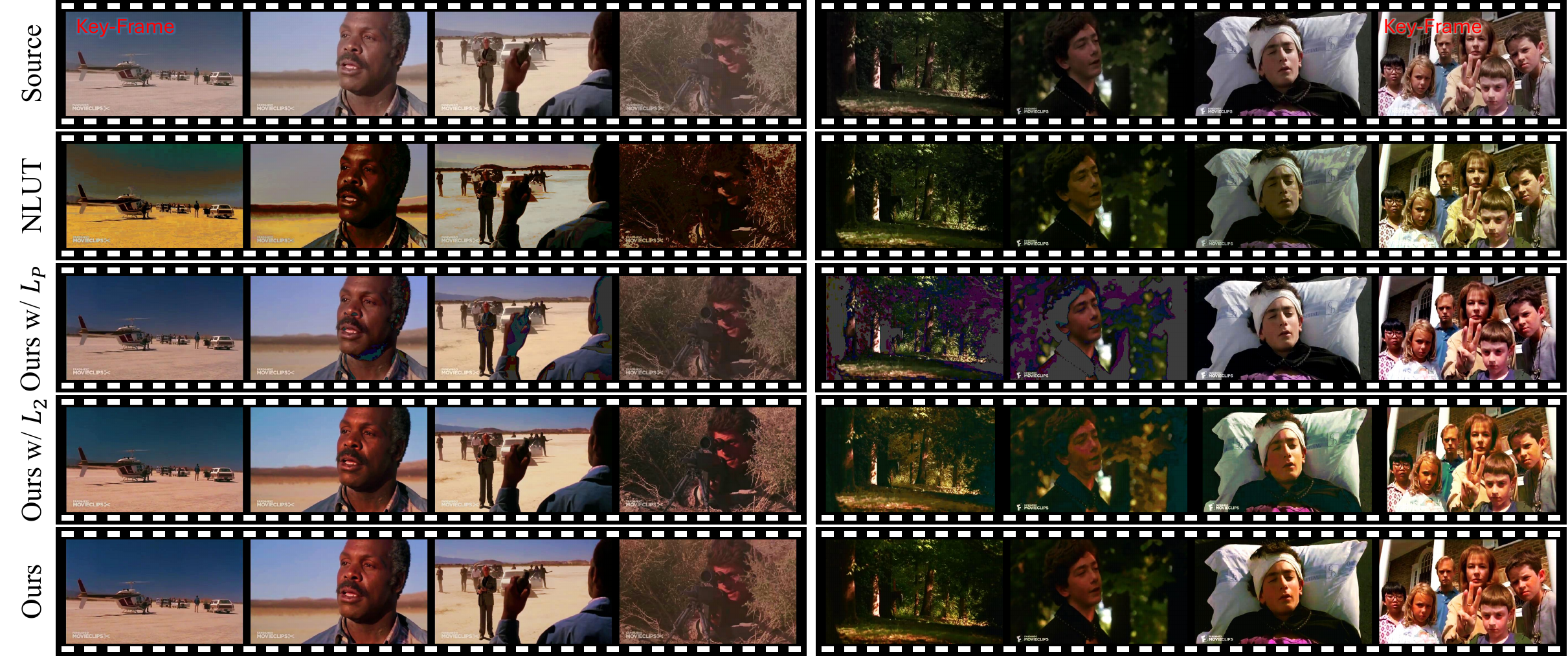}
    \vspace{-7.5mm}
    \caption{Comparison results of analysis. $\textit{L}_P$ and $\textit{L}_2$ denote perceptual loss and $\ell_2$ loss, respectively. (Please zoom-in)}
    \vspace{-5mm}
    \label{fig:frames}
\end{figure*}

\subsection{Quantitative Comparison}
\label{subsec:quantitative}
In this evaluation, we focus on measuring video quality both with and without its ground truth (GT). We use common quantitative metrics, PSNR, SSIM~\cite{wang2004image}, and LPIPS~\cite{zhang2018unreasonable}. 
To evaluate the quality of the video without reference, we use BRISQUE~\cite{mittal2011blind} and Blur Metric~\cite{crete2007blur}. BRISQUE evaluates the degree of image distortion using a model trained on pre-distorted images. Blur metric~\cite{crete2007blur} calculates the intensity and variance of neighboring pixels by applying a low-pass filter. Here, we do not measure temporal loss based on optical flow. Although it is widely used to evaluate video temporal consistency in style transfer tasks~\cite{wu2022ccpl, wen2023cap}, inaccurate temporal correspondences from the estimated optical flow make the evaluation metric unreliable~\cite{xu2021frame}.
Additionally, we compare the average inference time for 480 frames required to process a single video for each method. For a fair comparison, our inference time includes the key-frame selection step. Since NLUT includes a test time fine-tuning process, it is also considered in the result.

In \cref{tab:main_quant}, our method outperforms the comparison methods across most of the metrics in both datasets. It shows that the output images and videos from our method accurately resemble the reference frame while preserving the content integrity of the original frame. Furthermore, the shortest inference time achieved by our method is one of the advantages to realize practical applications in the movie industry.

The comparative methods, despite their impressive color transfer results, show unsatisfactory results due to the following reasons: (1) since they focus on adjusting overall color distributions between the reference and input videos, they cannot account for subtle tonal adjustments compared to ours. (2) feature transformation methods~\cite{chiu2022photowct2, an2020ultrafast, afifi2021histogan, wu2022ccpl, wen2023cap} directly generate the outputs, which inevitably leads to the loss of structural details when aligning feature distributions on latent spaces. Deep-Preset often fails to alter the color style as it can only handle basic color adjustments. Meanwhile, NLUT leads to noticeable artifacts in frames with dissimilar color distributions.

\subsection{Analysis}
\label{subsec:anal}

We first analyze the effectiveness of our explicit LUT generation. To do this, we modify our method to generate LUTs as implicit representation. Instead of minimizing a difference between input and output noise, we generate LUTs, apply them to input frames, and optimize the perceptual~\cite{zhang2018unreasonable} or $\ell_2$ distances between output and reference/GT frames. The other settings remain unchanged. Since the implicit representations of LUTs are primarily biased toward color information of input and reference frames, they show the distortion artifacts in sequential frames as shown in~\cref{fig:frames} and \cref{tab:analysis}. Although ours with $\ell_2$ distance yields consistent outputs in scenes with little variance, they suffer from the same artifacts in dynamic scenes.

\begin{table}[t]
    \centering
    \footnotesize
    \resizebox{\linewidth}{!}{%
    \begin{tabular}{c|ccccc}
    \hline \hline
    Method & PSNR$\uparrow$ & SSIM$\uparrow$ & LPIPS$\downarrow$ & BRISQUE$\downarrow$ & Blur$\downarrow$ \\
    \hline
    Ours w/ perceptual loss & 20.87 & 0.7560 & \underline{0.2666} & 43.38 & \textbf{0.4141} \\ 
    Ours w/ $\ell_2$ Loss & \underline{22.17} & \underline{0.7857} & 0.2773 & 44.10 & 0.4205 \\\hline
    NLUT w/ Explicit LUT & 13.97 & 0.5936 & 0.4679 & \underline{41.02} & 0.4244 \\
    Ours + LUT Bases-based &20.50&0.7321&0.3562&46.12& 0.4220\\
    Ours + MLP-based &19.73&0.7219  &0.4014&42.35&0.4344\\
    \hline
    
    \textbf{Ours} & \textbf{24.55} & \textbf{0.8445} & \textbf{0.1457} & \textbf{40.23} & \underline{0.4183} \\
    \hline \hline
    \end{tabular}
    }%
    \vspace{-3mm}
    
    \caption{Quantitative results of analysis.}
    \vspace{-7mm}
    
    \label{tab:analysis}
\end{table}

Next, we validate the advantage of our diffusion-based explicit LUT generation. We optimize the previous LUT generation, NLUT, by directly computing $\ell_2$ distance between output and GT LUTs. Additionally, we conduct experiments by replacing L-Diffuser with LUT bases-based~\cite{chen2023nlut}\,/\,MLP-based~\cite{lin2023adacm} LUT generations. 

In~\cref{tab:analysis}, all of them yield unsatisfactory results. In particular, the explicit version of NLUT performs worse than its original one. The key reason is that the deterministic method either lacks generative power or is prone to overfitting to a specific training set. Furthermore, LUT bases-based methods cannot cover all possible color styles with a finite set of bases, requiring additional fine-tuning. Note that it is impossible in this experiment due to the absence of GT LUTs at inference time. The MLP-based method treats each color value independently, making it difficult to achieve smooth, globally-consistent transformations.

As well-known, diffusion-based approaches are able to capture intricate data distributions and produce high-fidelity outputs conditioned on specific inputs~\cite{rombach2022high, zhang2023adding}.
Our method also leverages this capability to generate LUTs tailored to both diverse color attributes and semantic grading styles.

\begin{table}[t]
    \centering
    \footnotesize
    \resizebox{\linewidth}{!}{%
    \begin{tabular}{c|ccccc}
    \hline \hline
    Method & PSNR$\uparrow$ & SSIM$\uparrow$ & LPIPS$\downarrow$ & BRISQUE$\downarrow$ & Blur$\downarrow$ \\
    \hline
    Histogram + L-Diffuser & 18.52 & 0.6938 & 0.3877 & 41.96 & 0.4384 \\
    VGG + L-Diffuser & 18.95 & 0.7149 & 0.3165 & 41.16 & 0.4329\\
    Ours w/ Image Editor & 18.05 & 0.6413 & 0.4053 & 53.39 & 0.5278 \\ \hline
    Random&23.63&0.8169&0.1638&42.63& 0.4203\\
    Resnext&\underline{24.54}&0.8403  &\textbf{0.1439}&42.21&\underline{0.4189}\\
    DinoV2&24.46&\textbf{0.8486}&\underline{0.1442}&\underline{40.50}&\underline{0.4189}\\ \hline
    \textbf{Ours(CLIP)} & \textbf{24.55} & \underline{0.8445} & 0.1457 & \textbf{40.23} & \textbf{0.4183} \\
    \hline \hline
    \end{tabular}
    }%
    \vspace{-3mm}
    
    \caption{Quantitative results of our ablation study.}
    \vspace{-6.8mm}
    
    \label{tab:ablation}
\end{table}

\begin{figure*} [t]
    \centering
    \includegraphics[clip, width=\linewidth]{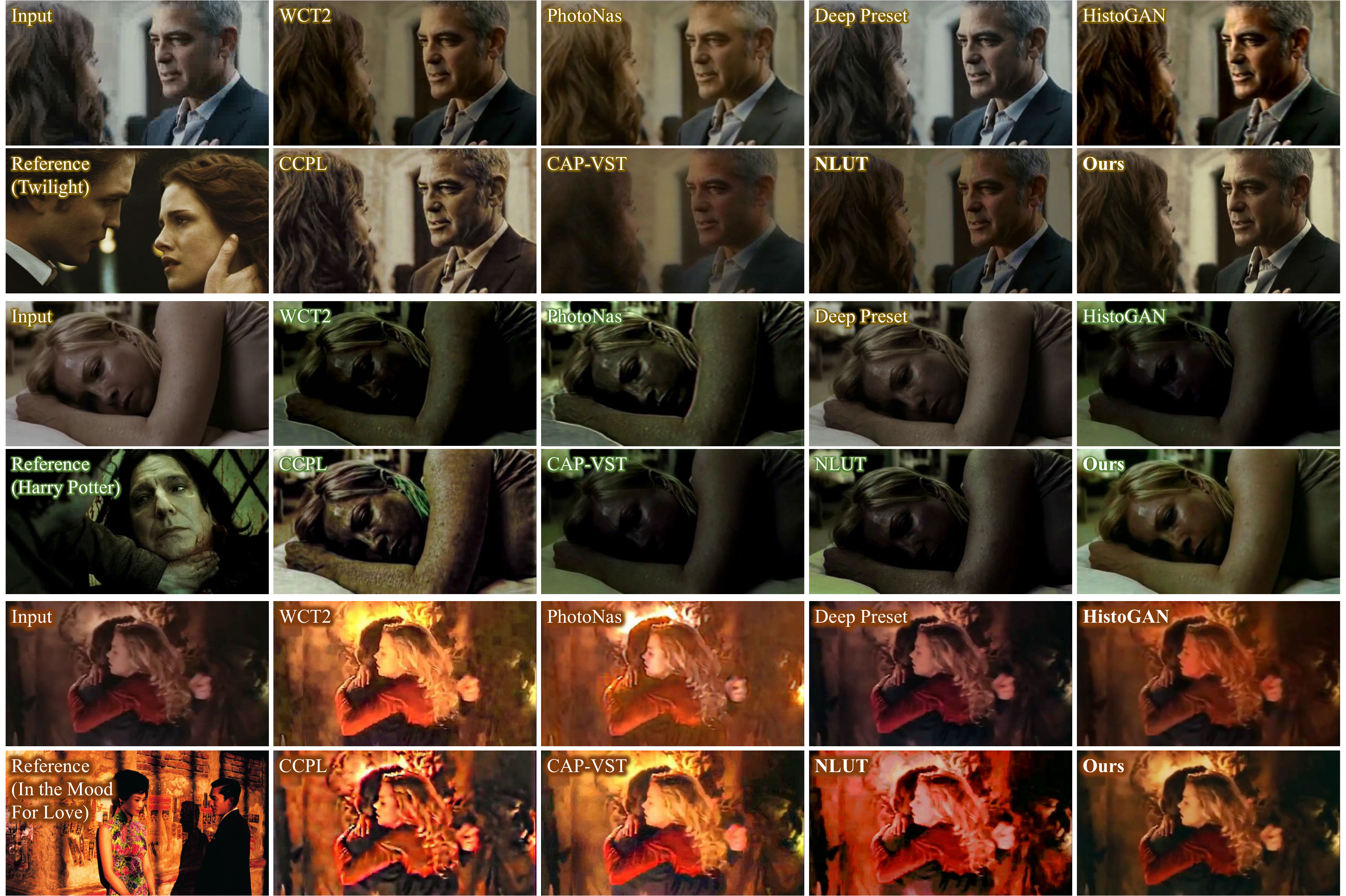}
    \vspace{-7mm}
    \caption{Example questionnaires for User Study 1, and outcomes from ours and the comparison methods}
    \label{fig:main_qual}
    \vspace{-5mm}
\end{figure*}
\vspace{-1mm}
\subsection{Ablation Study}
\label{subsec:abl}
We provide an ablation study to evaluate the functionality of each component in our method, GS-Extractor, L-Diffuser and CLIP for the key-frame selection. To validate the ability of GS-Extractor to capture high-level color features, we compare it to the color histogram vector from HistoGAN~\cite{afifi2021histogan} and the VGG model, both of which are commonly used in low-level color matching tasks. In this comparison, we replace GS-Extractor with them and use their features as conditioning inputs for L-Diffuser. Since each approach has a different feature dimensionality than our network, we include an additional MLP layer to adjust the feature dimensionality. We use the codes and pre-trained weights provided by the authors and fix these weights while training L-Diffuser and the MLP layer as described in \cref{subsec:training strategy}. 

As shown in \cref{tab:ablation}, the non-reference video quality metrics for them are nearly as good as ours, probably because these methods also generate LUTs that help prevent structural loss of the original content. However, they perform suboptimally on other metrics. Both alternatives fail to accurately capture the color characteristics of the reference frame. Since color histograms are inherently limited to global and aggregated color information, they can only capture color distributions without contextual color relationships. The VGG features are relatively useful for representing scene textures or patterns rather than color features, as discussed in many previous studies~\cite{hermann2020origins, ritter2017cognitive}.

In addition, to validate the effectiveness of L-Diffuser in preserving structural information, we compare it to the image editor used to train GS-Extractor in~\cref{subsec:overview}. Since we already have the weights of the image editor, we use them as fixed values while adding and training only the temporal blocks.
Despite good color representation in the image editor, the final output shows blurry artifacts in high-frequency regions. This is because direct generation of video frames often suffers from structural loss caused by the compression-and-reconstruction processes in latent diffusion frameworks~\cite{rombach2022high}. In contrast, our method preserves the original content by taking advantage of the generated LUT, achieving both color accuracy and structural fidelity.

Finally, we replace CLIP with other image embedding models~\cite{oquab2023dinov2, xie2017aggregated} for key-frame selection, following the same experimental setup in~\cref{subsec:setup}. In~\cref{tab:ablation}, the choice of image embedding model does not significantly impact on the performance. However, when key-frame pairs are selected randomly, the performance degrades considerably, highlighting the importance of the key-frame selection step. Additionally, CLIP features are used in the cross-attention module. Therefore, re-using them for the key-frame selection is a reasonable choice with respect to the overall computation reduction.

\subsection{User Study}
\label{subsec:userstudy}
To examine video qualities and subtle details that are difficult to fully capture through numerical metrics, we conduct an extensive user study via Amazon M-Turk~\cite{crowston2012amazon} following prior works~\cite{shin2024close, park2024kinetic}. Our study consists of two parts: (1)~\textbf{Study 1}: Comparison with other methods; (2)~\textbf{Study 2}: Validation of our pipeline. We conduct each study in separate groups because each group can influence the results of the other, which can lower the reliability of user studies. Each study has $15$ questionnaires with $20$ participants. The questionnaires are composed of $15$ input videos from our movie test set and collect $15$ reference frames from famous movies with their own iconic color grading effects~\cite{iconic}. When composing the input video for each reference frame, we impose a constraint that they must belong to the same genre. This is based on the best practice that users should at least use references from movies of the same genre when providing examples for color grading~\cite{vebrianto2023exploratory, colorgrading2024}.
In addition, considering that a lack of understanding the concept of color grading might affect the reliability of the study, we add a preliminary session to introduce the color grading. We summarize both technical and aesthetic purposes of video color grading. At the end of the session, we show some examples of movies to participants before and after colorists' work.

\noindent\textbf{Study 1.} This study is designed to compare our method with the comparison methods. Example questionnaires are shown in~\cref{fig:main_qual}. 
The study consists of three different parts. First, we compare the qualities of the output videos of each method. Users are asked to select all videos that have artifacts such as blur, color bleeding, etc. To eliminate potential violations, such as participants randomly choosing answers, we include some dummy choices in certain questions.

\begin{table}[t]
    \centering
    \resizebox{\linewidth}{!}{%
    \begin{tabular}{c|c|c|c}
    \hline \hline
     Method & Artifact $\downarrow$ & Mood Matching $\uparrow$ & Intent Matching $\uparrow$ \\
    \hline
    WCT2 & 0.38 & \underline{3.80}~$\pm$~1.35& 3.62~$\pm$~1.28 \\
    PhotoNas & 0.33 & 3.43~$\pm$~1.36 & 3.62~$\pm$~1.39 \\
    ~~~Deep Preset~~~ & \underline{0.17} & 3.64~$\pm$~1.53 & 3.53~$\pm$~1.36 \\
    HistoGAN & 0.57 & 3.43~$\pm$~0.88 & \underline{3.76}~$\pm$~0.78 \\
    CCPL & 0.56 & 3.11~$\pm$~0.95& 3.27~$\pm$~1.01 \\
    CAP-VST & 0.42 & 3.63~$\pm$~0.90 & 3.46~$\pm$~0.97 \\
    NLUT & 0.41 & 3.69~$\pm$~0.93 & 3.66~$\pm$~0.79 \\
    \hline
    Ours & \textbf{0.16}& \textbf{4.14}~$\pm$~1.06&  \textbf{3.97}~$\pm$~1.03 \\
    \hline \hline
    \end{tabular}
    }
    \vspace{-3mm}
    \caption{Comparison results of User Study 1.}
    \label{tab:user_study}
    \vspace{-6mm}
\end{table}

In the following part, participants are first asked to watch a reference frame for a given amount of time to \textit{grasp} the mood, emotion, and intent of the colorist. They are then presented with a series of videos produced by ours and the comparison methods, and asked to rank each one based on how effectively it conveys what they believe to be the colorist's mood, emotion, and intent. Participants rank the videos from 1 (most closely matching) to 8 (least closely matching), and these rankings are converted into scores in reciprocal order, ranging from 7 to 0.

In the final part of this study, we evaluate how well each model reflects the \textit{true intent of the colorist} who retouches the reference frame. We collect analyses for reference frames from~\cite{iconic} and provide it to the participants as a description along with the reference frames. Similar to the previous part, they watch eight videos in random order and rate them based on how well they match the description.

The performance of each method is evaluated based on the selection rate in the first part and the participant scores in the remaining parts. As shown in~\cref{tab:user_study}, our approach achieves the most promising performance in all studies, supporting its ability to effectively transfer key mise-en-scène features such as emotion and mood, which are crucial factors in video color grading, while preserving the integrity of the original content. 

\begin{figure} [t]
    \centering
    \includegraphics[clip, width=\linewidth]{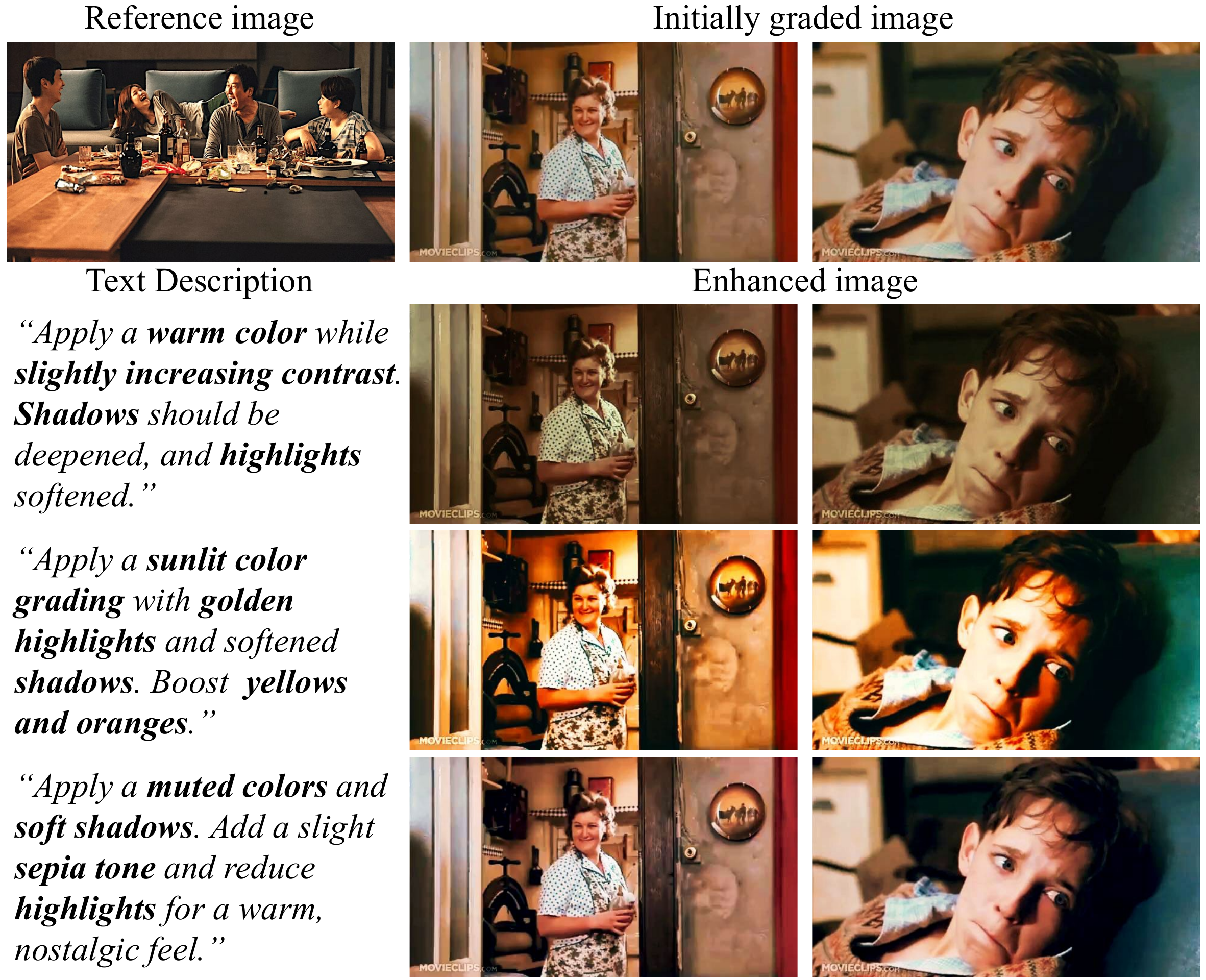}
    \vspace{-6mm}
    \caption{Examples of user feedback on initally graded image.}
    \vspace{-5mm}
    \label{fig:user_feedback}
\end{figure}

\noindent\textbf{Study 2.} Next, we demonstrate the validity of our pipeline for incorporating user feedback in this study. Since real-time interaction with participants is challenging, we first generate $10$ candidate prompts that users can select to modify the output videos accordingly. Participants first view the reference frame along with the input and output videos generated by our method. They are then presented with options for further modification, including the option of no further adjustment. Based on their choices, they then compare the original output videos with the further modified versions of their choices. For those who select an option other than ``no further adjustment'', a scoring system is used where they rate from 1 (indicating the original video is better) to 5 (indicating the modified video is better). Examples of the questionnaires used in this study are shown in~\cref{fig:user_feedback}.

The average score of the results is $3.64$ with a standard deviation of $1.18$. Since a score of $3$ represents a balance between the original and modified videos, the higher score than $3$ indicates that participants prefer the modified videos, even with limited modification choices. This result supports the effectiveness of our additional retouching feature.

\section{Conclusion}

We present a novel video color grading approach. Going beyond simple color distribution matching, we focus on color grading based on the mise-en-scène of a reference image/video and incorporating user feedback to fully accomplish the video color grading task. To do this, we explicitly predict a LUT with a diffusion model conditioned on the high-level feature differences between the input and reference frames. In addition, our framework allows the user to express their preferences through textual prompts. Extensive quantitative experiments and user studies show that our method outperforms existing approaches.

\vspace{3mm}

\noindent\textbf{Limitations \& Future Work}
There is still rooms for improvements. As shown in~\cref{fig:frames}, our method produces consistent outputs even across different shots. However, for occurrence of scene changes, our current pipeline will face challenges. Its solution to such scenarios would be interesting directions as future work. We believe that this issue can be tackled by incorporating techniques like video scene segmentation~\cite{tan2024neighbor, wu2022scene}, where a unique LUT is generated for each scene segment. In addition, our framework does not support spatially-variant color grading, which transfers localized features based on semantic correlations between the reference and source images. We expect that incorporating semantic features~\cite{kirillov2023segment, ravi2024sam} or structure priors~\cite{shin2023task, nazeri2019edgeconnect} could be promising solutions to address this challenge.

\vspace{1mm}

\noindent\textbf{Acknowledgment}
This work was supported by the National Research Foundation of Korea(NRF) grant funded by the Korea government(MSIT)(RS-2024-00338439), `Project for Science and Technology Opens the Future of the Region' program through the INNOPOLIS FOUNDATION funded by Ministry of Science and ICT (Project Number: 2022-DD-UP-0312) and Information $\&$ Communications Technology Planning $\&$ Evaluation (IITP) grant funded by the Korea government(MSIT) (No.RS-2025-25441838, Development of a human foundation model for human-centric universal artificial intelligence and training of personnel).

{
    \small
    \bibliographystyle{ieeenat_fullname}
    \bibliography{main}
}

\end{document}